\pdfoutput=1
\documentclass[11pt]{article}

\usepackage[]{acl}

\usepackage{times}
\usepackage{latexsym}
\usepackage{amsmath}
\usepackage{amssymb}
\usepackage{graphicx}
\usepackage{multirow}
\usepackage{makecell}
\usepackage{float}
\usepackage{bbding}
\usepackage[linesnumbered,ruled,vlined]{algorithm2e}

\usepackage[T1]{fontenc}
\usepackage[utf8]{inputenc}

\usepackage{microtype}
\usepackage{amsmath}
\usepackage{graphicx}

\title{In-Context Learning with Reinforcement Learning for Incomplete Utterance Rewriting}

\author{Haowei Du,    Dongyan Zhao}
\begin{document}
\maketitle
\begin{abstract}
In-context learning (ICL) of large language models (LLMs) has attracted increasing attention in the community where LLMs
make predictions only based on instructions augmented
with a few examples. 
Existing example selection methods for ICL utilize sparse or dense retrievers and derive effective performance. However, these methods do not utilize direct feedback of LLM to train the retriever and the examples selected can not necessarily improve the analogy ability of LLM. To tackle this, we propose our policy-based reinforcement learning framework for example selection (RLS), which consists of a language model (LM) selector and an LLM generator. The LM selector encodes the candidate examples into dense representations and selects the top-k examples into the demonstration for LLM. The outputs of LLM are adopted to compute the reward and policy gradient to optimize the LM selector. We conduct experiments on different datasets and significantly outperform existing example selection methods. Moreover, our approach shows advantages over supervised finetuning (SFT) models in few shot setting. Further experiments show the balance of abundance and the similarity with the test case of examples is important for ICL performance of LLM.
\end{abstract}

\section{Introduction}

\begin{table}[t]
\centering
\begin{tabular}{l|c}
\hline
 & \textbf{Utterance}\\
\hline
\multirow{2}{*}{$u_1$}& Hello, I am looking for an expensive \\
& restaurant that serves fusion food. \\
\hline
\multirow{2}{*}{$u_2$} &  I 'm sorry, there are no fusion restaurants \\
&  listed in the expensive price range. Would\\
&  you like to try something else?\\ 
\hline
$u_3$ &How about Mediterranean food?   \\
\hline
\multirow{2}{*}{$u_3^*$} & How about Mediterranean food \\
 & in expensive price range?\\
\hline
$e_1$ & How about serving Mediterranean food?\\
\hline
\multirow{2}{*}{$e_1^*$} & How about a cheap restaurant\\
& serving Mediterranean food? \\
\hline
\multirow{2}{*}{$e_2$} & Can you recommend a restaurant to me? 
\\ & I don't want to spend a lot of money.\\
\hline
\multirow{3}{*}{$e_2^*$} & Can you recommend a restaurant to me \\
& in the south part of town? I don't\\
& want to spend a lot of money. \\
\hline
\end{tabular}
\caption{One example from CANARD dataset. $u_1$-$u_2$ are 2 turns of contextual utterances, $u_3$ denotes the incomplete utterance, $u_3^*$ denote the golden rewritten utterance, $e_1/e_1^*$ and $e_2/e_2^*$ denote two candidate example pairs of incomplete utterance and rewritten utterance which can be prompted to the LLM. }
\label{case 1}
\vspace{2em}
\begin{tabular}{l|c|c|c}
\hline
\textbf{Example} & \textbf{Sparse} & \textbf{Dense} & \textbf{ROUGE}\\
\hline
$e_1$ & \Checkmark  & \Checkmark   & 50.0\\
\hline
$e_2$ & \XSolidBrush  & \XSolidBrush  & 55.6\\
\hline
\end{tabular}
\caption{The Rouge score of ChatGLM by use of different examples in prompts. ``Sparse'' denotes selecting the example by sparse retrieval methods like BM25, ``Dense'' denotes selecting the example by dense representations by PTM.}
\label{selection}
\end{table}

In recent years, there has been a growing focus on multi-turn dialogue modeling \cite{choi2018quac, sun2019dream, reddy2019coqa}. One of the primary challenges in this field is the tendency of speakers to employ incomplete utterances, such as co-reference or ellipsis, when referring back to entities or concepts that have been mentioned in the dialogue history. \citet{su2019improving} demonstrates that ellipsis and co-reference occur in over 70\% of dialogue utterances. To address this phenomenon, the Incomplete Utterance Rewriting (IUR) task, as proposed by \cite{pan2019improving, elgohary2019can}, aims to rephrase an incomplete utterance into a self-contained utterance that is semantically equivalent and can be comprehended independently, without relying on contextual information.

Recent generation-based approaches for IUR tackle this task as a seq2seq problem and gain impressive performances \cite{pan2019improving, huang2021sarg, inoue2022enhance}. 
With the increasing ability of LLMs, ICL
has become a new direction for natural language generation, where LLMs
make predictions only depending on contexts augmented
by a few examples (demonstration) without weights updating \cite{brown2020language,chowdhery2022palm,touvron2023llama}. However, the performance of ICL is sensitive to the
selection of in-context examples \cite{zhao2021calibrate,liu2021makes,dong2022survey}.

We take a case from TASK dataset in Table \ref{case 1} and \ref{selection}, where the incomplete utterance is ``How about Mediterranean food?'' and the omitted part is the postpositive attributive ``in expensive price range''. By metrics of sparse retrieval methods like BM25 \cite{robertson2009probabilistic} or dense retrieval methods like PTM \cite{liu2022few}, the example incomplete utterance $e_1$ as well as its contexts and rewritten utterance should be selected to be in-context examples. However, the performance of LLM for this question with example $e_1$ drops by 5 ROUGE score compared with example $e_2$.

To directly select the examples that can improve the analogy ability of LLM, we introduce policy based reinforcement learning (RL) into example selection for IUR (RLS). Given a set of candidate examples, we utilize a small-scale language model (LM) to encode the context and incomplete utterance of each example. Intuitively, if a subset of candidate examples leads to increasing performance of LLM for IUR, RLS should
assign high scores to them; the more the performance
increases, the higher the example scores should be. Therefore, we compute the reward by ICL performance and optimize the LM selector with policy gradient.

We conduct experiments on three benchmark datasets in IUR field and compare with existing competitive example selection methods. Our approach outperforms existing methods by about 2 score in CANARD and REWRITE dataset and 10 score in TASK dataset with different metrics including BLEU, ROUGE and F-score.

Our contributions can be summarized as:\\
\textbf{1.} We are the first to explore ICL performance of LLM for IUR task and design the effective formulation of demonstration for IUR task.\\
\textbf{2.} We introduce policy-based RL into example selection for ICL prompts, which directly utilize the LLM feedback to train the LM selector.\\
\textbf{3.} Our approach significantly outperforms existing example selection methods across sparse retrieval and dense retrieval methods. \\
\textbf{4.} Our approach shows advantages against SFT models in few shot setting. We explain the improvement comes from the linguistic complexity and abundance, as well as the similarity with the test case of examples.

\begin{figure*}
\centering
\includegraphics[width=0.8\textwidth]{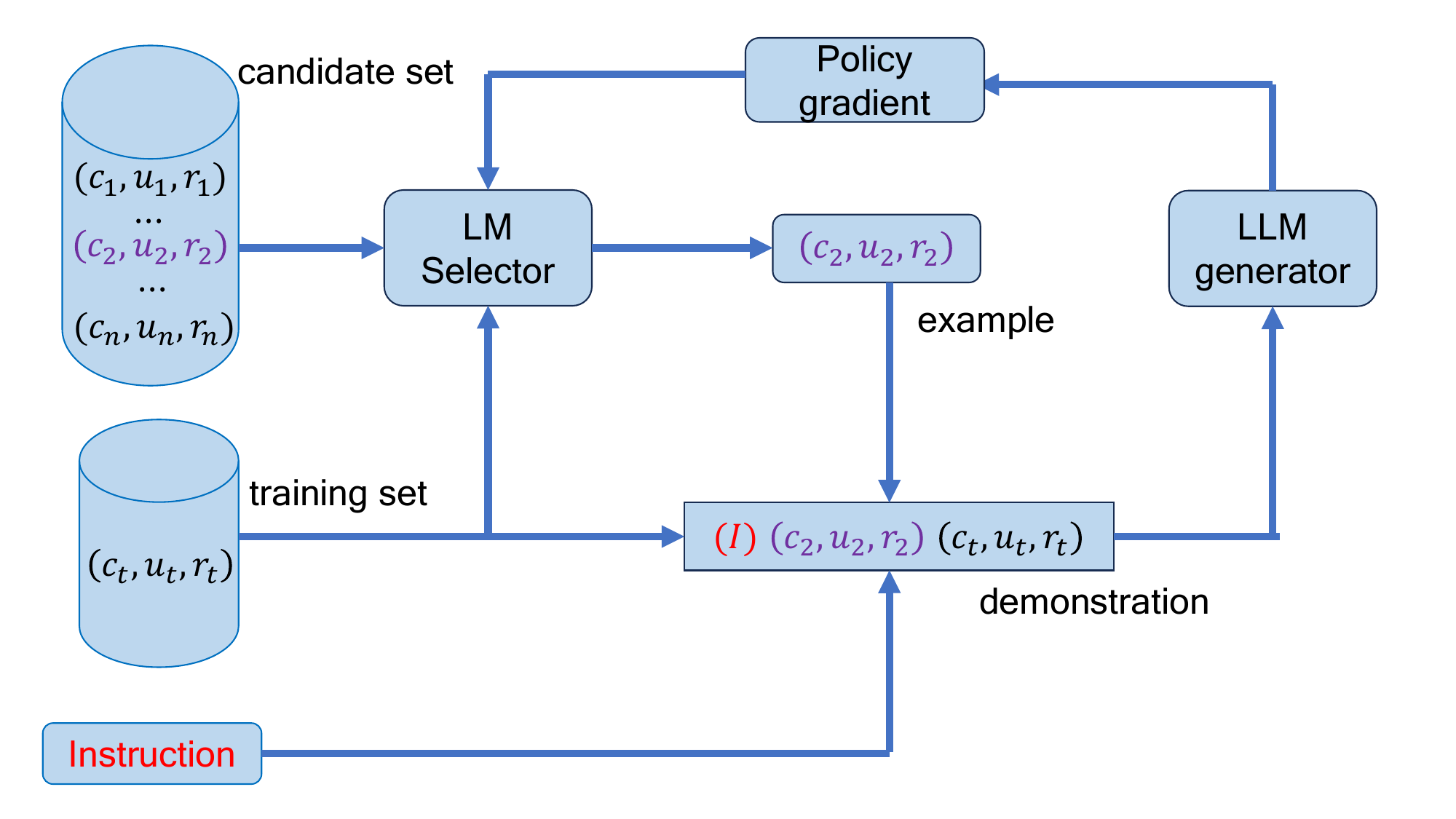}
\caption{Method Overview. Our approach consists of an LM example selector and an LLM generator. }
\label{pipeline}
\end{figure*}

\section{Related Work}
There are two main streams of approaches to tackle the task of IUR: edit-based and generation-based. Generation-based models solve this task as a seq2seq problem, which is more relevant to our approach for ICL with LLM. \citet{su2019improving} utilize pointer network to respectively predict the prob of tokens in rewritten utterance from contexts or incomplete utterance. 
\citet{hao2021rast} formulate the task as sequence tagging to reduce the search space.
\citet{huang2021sarg} combine a source sequence tagger with
an LSTM-based decoder to maintain grammatical correctness. 

Demonstration selection is crucial to ICL and \citet{liu2021makes} showed that downstream performance can
vary widely depending on the choice of in-context
examples. 
\citet{liu2022few} utilize sentence representations of PTM to select the examples with more cosine similarity. \citet{sorensen2022information} and \citet{gonen2022demystifying} argue that mutual information and perplexity are also valuable selection metrics which do not need labeled examples and specific LLM. 
\citet{levy2022diverse} select diverse demonstrations to collectively cover all of the structures required
in the outputs.
\citet{kim2022self} generate
demonstrations for ICL from
PLM itself to minimize the reliance on the
external demonstration.
\citet{rubin2021learning} first build an unsupervised
retriever like BM25 to recall similar
examples as candidates and then construct a supervised
retriever to select demonstrations from candidates
However, these methods fail to directly select the examples into demonstrations that can improve the analogy ability of LLM for IUR task and some useful examples like in table \ref{case 1} will be neglected.

 % \IncMargin{1em}
% \begin{algorithm} \SetKwData{Left}{left}\SetKwData{This}{this}\SetKwData{Up}{up} \SetKwFunction{Union}{Union}\SetKwFunction{FindCompress}{FindCompress} \SetKwInOut{Input}{input}\SetKwInOut{Output}{output}
	
% 	\Input{PTM $f_m^{\theta_0}$, LLM $f_M$, candidate example set $C$, training set $T$, validation set $V$, learning rate $\alpha$, number of selected examples $K$, maximum iteration times $N$, inner iteration times $N_0$} 
% 	\Output{The scorer policy model $f_m^{\theta_*}$}
% 	 \BlankLine 
%   \emph{Store $\theta \leftarrow \theta_0$} \;
% 	 \For{$i\leftarrow 1$ \KwTo $N$}{ 
% 	 	\For{$j\leftarrow 1$ \KwTo $N_0$}{
%    sample a batch of cases from $T$ \;
%     compute the representations and score the examples in $C$ with the current training case $T_j$ by the policy $f_m^{\theta}$\;
%     sample $K$ examples by normalizing the score as prob \;
%     formulate the demonstration and input to $f_M$ to derive the rewritten utterance $r_j$ \;
%     evaluate $r_j$ with golden label to compute the reward for the current training case $R(a_j, s_j)$ \;
%     $\theta \leftarrow  \theta + \alpha \nabla_{\theta} E_{s \sim p_{\theta}(s)} R(s)$
%     }
%     sample a batch of cases from $V$, evaluate the examples selected  by $f_m^{\theta}$ to decide if interrupting the iteration.}
%  	 	  \caption{RLS}
%  	 	  \label{algo} 
%  	 \end{algorithm}
 % \DecMargin{1em} 
 
\section{Task Definition}
\label{format}
Given the context $C$ and the incomplete utterance $U$, we aim to derive the rewritten version $R$ which can be comprehended without the context. The IUR task is to learn a map function $f(C, U|\theta) = R$. ICL with LLM for IUR task needs a set of candidate examples $\{y_1, y_2,\cdots, y_N\}$, where $y_i = (C_i, U_i, R_i), 1\leq i \leq n$ denotes the i-th example and $N$ denotes the number of candidate examples. For a test case $x = (c_x, u_x)$, we select $k$ examples from the candidate set
and the input to LLM $M$ is the concatenation of the task instruction $I$, the $k$ demonstration examples, and the test case $x$. The rewritten utterance of $x$ is generated by $M$: $R_x = M[I; (C_{i_1}, U_{i_1}, R_{i_1}); (C_{i_2}, U_{i_2}, R_{i_2}); \cdots; \\(C_{i_k}, U_{i_k}, R_{i_k}); (C_{i_x}, U_{i_x})]$, $;$ denotes concatenation of texts. So the key of ICL to solve IUR task is to select the appropriate examples.

\section{Methodology}
% By fig. \ref{pipeline}, our approach consists a LM selector and a LLM generator, where the selector selects examples into the demonstration for LLM, and the rewards computed by generation from LLM are utilized to optimize the selector by policy based RL.
% Our objective is to achieve the highest possible accuracy on unseen test examples by receiving up to K annotations from a set of unlabeled examples. However, the space of potential prompts increases exponentially with the number of unlabeled examples, making it impossible to list them all. Therefore,
We approach the example selection as a sequential decision-making problem. 
State space: the sequence of selected demonstration examples $y_{i_1}, y_{i_2}, \cdots, y_{i_k}$, where $k$ denotes the number of selected examples.
Action space: select the next example given the current selection result and insert the concatenation of its contexts, incomplete utterance and rewritten utterance into demonstrations.
Policy: We utilize transformer architecture to model the probability of each example in the candidate set to be selected into the demonstration. 
\subsection{Utterance Encoder}
We make use of BERT \cite{devlin2018bert} to encode the semantic and syntactic information of utterances. For each example in the candidate set and the current test case, the input to PTM is the concatenation of contexts and incomplete utterance, which is separated by ``[SEP]'' token. The hidden state corresponding to ``[CLS]'' token is used to represent the case.
\begin{align}
    \mathbf{S_i} &= \mathbf{BERT}([CLS; c_i, SEP, u_i])  \\
    \mathbf{S_x} &= \mathbf{BERT}([CLS; c_x, SEP, u_x]
\end{align}
where $c_i$ and $u_i$ as well as $c_x$ and $u_x$ denote the context and the incomplete utterance of i-th candidate example and the current test case. To be consistent with the inference, we do not include the rewritten utterance into representing each case.
\subsection{Scoring Policy} \label{policy}
With the representation of each case, we model the probability of each candidate case to be selected as in-context examples by metrics of cosine similarity:
\begin{align}
    p_i &= \frac{exp(e_i)}{\sum_{1\leq j \leq N} exp(e_j)} \label{prob} \\
    e_{i} &= \frac{|\mathbf{S_i W S_x}|}{\Vert \mathbf{S_i} \Vert * \Vert \mathbf{S_x} \Vert} 
\end{align}
where $\Vert .\Vert$ denotes L2 norm, $N$ denotes the number of candidate examples. We sample $k$ demonstration examples according to the probability computed.

\begin{table}
\centering
\begin{tabular}{l|c|c|c}
\hline
 & CANARD & TASK & REWRITE \\
\hline
Language & English & English & Chinese \\
\# Train  & 32K & 2.2K & 18K \\
\# Dev &  4K & 0.5K &2K \\
\# Test  &  6K & NA & NA \\
Con. len  &  85.4 & 52.6  & 17.7 \\
Cur. len  & 7.5 & 9.4 & 6.5 \\ 
Rew. len   & 11.6 & 11.3 & 10.5  \\
\hline
\end{tabular}
\caption{ Statistics of datasets. ``Con. len'', ``Cur. len'', ``Rew. len'' denote the length of contexts, incomplete utterance and rewritten utterance respectively.}
\label{stat}
\end{table}

\subsection{Policy Iterating}
Intuitively, if a subset of candidate examples leads to increasing performance of LLM to rewrite the current incomplete utterance, RLS should assign high scores to them; the more the performance
increases, the higher the example scores should be. For each case in training set, we input the demonstration formulated as section \ref{format} containing the examples selected by the current policy into the LLM. The performance metrics of outputs from the LLM against the golden rewritten utterance are utilized as the rewards of the policy.

To maximum the expected rewards of the current policy, we compute the gradient of policy as follows:
\begin{align*}
    & \nabla E_{s \sim p(s)} R(s) \\
    &= \int \nabla (p(s) R(s)) ds\\
    & = \int p(s) \nabla (log(p(s))R(s))ds \\
    & = E_{s\sim p(s)} \nabla log(p(s))R(s) \\
    & = E_{s\sim p(s)} \sum_{t=1}^K \nabla log(p(a_t|s_t))(\sum_{t=1}^K R(a_t,s_t)) \nonumber \\
    & = \frac{1}{n} \sum_{i=1}^n \sum_{t=1}^K \nabla log(p(a_{i,t}|s_{i,t}))(\sum_{t=1}^K R(a_{i,t},s_{i,t})) \nonumber 
\end{align*}
where $K$ denotes the number of demonstration examples, $n$ denote the size of training set and $R(a_{i,t},s_{i,t})$ denotes the reward computed by inputing the current examples selected $s_{i,t}$ to LLM for i-th case in training set.

To reduce the generation time cost of LLM, we simplify the computation of $\sum_{t=1}^K R(a_{i,t},s_{i,t})$ by replacing with $R(a_{i, K}, s_{i, K})$. That is, in the process of selecting examples, we do not input the intermediate demonstrations into LLM to derive the reward. Instead, after selecting all the K examples, we input the final demonstration to LLM. To reduce the variance of learning process, we modify the reward by subtracting a baseline score, which will not affect the gradient. In practice, we compute the ICL performance with randomly selecting examples as the baseline score for IUR task. So the gradient can be simplified as 
\begin{align*}
    & \nabla E_{s \sim p(s)} R(s) \\
    & = \frac{1}{n} \sum_{i=1}^n \sum_{t=1}^K \nabla log(p(a_{i,t}|s_{i,t}))R(a_{i,K},s_{i,K}) \\
     & = \frac{1}{n} \sum_{i=1}^n R(a_{i,K},s_{i,K}) \sum_{t=1}^K \nabla log(p(a_{i,t}|s_{i,t})) \\
\end{align*}

With the policy gradient computed, we optimize the parameters of policy model defined in section \ref{policy}. We compute the performance on validation set to control the termination of iterating process. In practice, we sample a small subset from the original training dataset to compose the candidate example set $C$ and another disjoint subset to form $T$ to train our RLS algorithm.
% The conclusion of our algorithm is in Algorithm. \ref{algo}.
% \subsection{Algorithm}

% We conclude the pipeline of our algorithm in alg. \ref{algo}. 
% In practical, we sample a small subset from the original training dataset to compose the candidate example set $C$ and another disjoint subset to form $T$ to train our RLS algorithm.

\begin{table*}
\centering
\begin{tabular}{l|c|c|c|c|c|c|c|c|c|c}
\hline
 \multirow{2}{*}{Model}  & \multicolumn{3}{c|}{ROUGE} & \multicolumn{4}{c|}{BLEU} & \multicolumn{3}{c}{F-score} \\
 \cline{2-11}
 &  \textbf{RL} &  \textbf{R1} & \textbf{R2} &  \textbf{B1} & \textbf{B2}  & \textbf{B3} & \textbf{B4} & \textbf{F1} & \textbf{F2}  & \textbf{F3}  \\
\hline
Random &  52.41 &  54.02 &  38.74 & 48.40 & 40.80 & 35.08 & 29.70 & 25.03 & 17.65 & 14.26   \\
BM25 & 53.92 & 56.16 & 40.14 & 51.03 & 43.05 & 37.03 &  31.44 & 30.00 & 20.19 & 15.62 \\
KATE & 53.00 & 55.24 & 39.61 & 49.31 & 41.60 & 35.82 & 30.47 & 28.99 & 19.51 & 15.23  \\
EPR & 54.08 & 56.26 & 40.23 & 51.59 & 43.54 & 37.51 & 31.91 & 29.59 & 19.96 & 15.62  \\ 
BSR & 54.28 & 56.39 & 40.50 & 51.99 & 44.00 & 38.01 &  32.44 & 29.61 & 20.15 & 15.90 \\
\hline
Ours & \textbf{55.69} & \textbf{57.55} & \textbf{42.22} & \textbf{53.29} & \textbf{45.55} & \textbf{39.65} & \textbf{34.05} & \textbf{29.52} & \textbf{20.59} & \textbf{16.46}  \\
\hline
\end{tabular}
\caption{ ICL Evaluations of ChatGLM with 5-shot demonstrations on CANARD dataset. Our approach significantly outperforms existing example selection methods, where p-values of ROUGE, BLEU and F-score are smaller
than 0.001.}
\label{canard}
\end{table*}

\begin{table*}
\centering
\begin{tabular}{l|c|c|c|c|c|c|c|c|c|c|c|c}
\hline
\multirow{3}{*}{Model} & \multicolumn{6}{c|}{TASK} & \multicolumn{6}{c}{REWRITE}\\
\cline{2-13}
 & \multicolumn{3}{c|}{ROUGE} & \multicolumn{3}{c|}{BLEU}  &\multicolumn{3}{c|}{ROUGE} & \multicolumn{3}{c}{BLEU} \\
 \cline{2-13}
 &  \textbf{RL} &  \textbf{R1} & \textbf{R2} &  \textbf{B1} & \textbf{B2}  & \textbf{B3} 
& \textbf{RL} &  \textbf{R1} & \textbf{R2} &  \textbf{B1} & \textbf{B2}  & \textbf{B3} \\
\hline
Random &  44.5 &  45.8 & 31.8 & 34.1 & 28.6 & 25.3 
& 64.4 &  66.8 & 54.0 & 61.2 & 55.3 & 49.7 
\\
BM25 & 46.2 & 47.5 & 34.1 &  33.1 & 27.9 & 24.8 
& 64.6 & 67.0 & 55.6 &  65.7 & 60.1 & 54.7 
\\
KATE & 45.8 & 47.3 & 34.4 & 33.1 & 28.0 & 25.0    
& 63.2 & 65.2 & 53.2 & 61.6 & 55.9 & 50.2 
\\

EPR & 49.2 & 50.4 & 37.0 & 35.4 & 30.1 & 27.0 
& 65.8 & 68.1 & 56.4 & 66.7 & 61.1 & 55.8  \\

BSR & 49.7 & 51.6 & 37.8 &   35.2 & 29.6 & 26.4
& 65.3 &  68.0 & 56.3 &  65.8 & 60.1 & 54.7 
\\
\hline
Ours & \textbf{57.2} & \textbf{57.9} & \textbf{45.1} & \textbf{43.0} & \textbf{38.3} & \textbf{35.2}
 & \textbf{66.5} & \textbf{68.3} &  \textbf{56.8} & \textbf{67.2} &  \textbf{61.7} & \textbf{56.4}  
 \\
\hline
\end{tabular}
\caption{ ICL Evaluations of ChatGLM with 5-shot demonstrations on Task and REWRITE dataset. Our approach significantly outperforms existing example selection methods, where p-values of ROUGE, BLEU and F-score are smaller
than 0.001.}
\end{table*}

\section{Experiments}
\subsection{Datasets and LLM}
Following \citet{liu2020incomplete}, \citet{inoue2022enhance} and \citet{zhang2022self} we conduct experiments on three benchmark datasets across different languages and domains in IUR field. CANARD dataset \cite{elgohary2019can} includes English conversational question answering; TASK dataset \cite{quan2019gecor} contains task-oriented English dialogues; REWRITE dataset \cite{su2019improving} is composed of open-domain Chinese dialogues.

We choose ChatGLM-6B\footnote{https://github.com/THUDM/ChatGLM-6B} as the LLM $f_M$ which does not conduct parameter updating. ChatGLM is a bilingual large language model pretrained by supervised finetuning, instruction tuning and human feedback reinforcement learning, which is suitable for our bilingual datasets.

\subsection{Metrics}
Following \citet{liu2020incomplete, inoue2022enhance, zhang2022self} , we utilize BLEU \cite{papineni2002bleu}, ROUGE \cite{lin2004rouge} and F-score to evaluate the IUR task.
BLEU and ROUGE focus on the overall quality of the rewritten utterance and F-score concentrate more on words from the context (important words) which are argued to be harder to copy \cite{pan2019improving, inoue2022enhance}.

\subsection{Baselines}
We compare our approaches with competitive example selection methods as follows:

\paragraph{Random} In this baseline, we randomly select the examples from candidate set to formulate the demonstration.

\paragraph{BM25} \cite{robertson2009probabilistic} In this baseline, first we utilize SpaCy NLP tools \cite{vasiliev2020natural} to stem words in the context and incomplete utterance of each case. Then BM25 method is adopted to estimate the relevance of examples to a given test case. The top-k relevant examples are selected to formulate the demonstration. It belongs to the sparse retrieval methods.

\paragraph{KATE} \citet{liu2022few} make use of SBERT \cite{reimers2019sentence} to select examples which are semantically similar to the test sample and build a kNN-based unsupervised retriever. It belongs to the dense retrieval methods.

\paragraph{EPR} \citet{rubin2021learning} assume the unsupervised retriever can act as the guide to the LM retriever and propose a two-stage approach. It first builds an unsupervised
retriever (e.g., BM25) to recall surface similar
examples as candidates and then constructs a supervised
retriever EPR to select demonstrations from candidates. It can be seen as a combination of sparse and dense retrieval methods.

\paragraph{BSR} \citet{gupta2023coverage}
propose a novel framework for
selecting sets of maximally informative demonstrations for the salient aspects of the test input,
e.g., reasoning patterns, entities, etc. Examples selected using this framework are informative about
the test input and help the LLM understand and
perform the task

However, these methods fail to directly utilize the feedback by LLM and the examples selected can not necessarily improve the analogy ability of LLM.

\begin{table*}
\centering
\begin{tabular}{l|c|c|c|c|c|c|c|c|c|c}
\hline
 \multirow{2}{*}{Model}  & \multicolumn{3}{c|}{ROUGE} & \multicolumn{4}{c|}{BLEU} & \multicolumn{3}{c}{F-score} \\
 \cline{2-11}
 &  \textbf{RL} &  \textbf{R1} & \textbf{R2} &  \textbf{B1} & \textbf{B2}  & \textbf{B3} & \textbf{B4} & \textbf{F1} & \textbf{F2}  & \textbf{F3}  \\
\hline
Ours (100-100) & 54.22 & 56.13 & 40.86 & 51.08 & 43.49 & 37.74 & 32.30 & 28.60 & 20.02 & 16.08  \\
Ours (100-500) & 55.08 & 56.51 & 41.81 & 51.77 & 44.62 & 39.06 & 33.73 & 27.17 & 20.18 & 16.59 \\
Ours (500-100) & 54.12 & 56.29 & 40.73 & 51.49 & 43.65 & 37.74 & 32.22 & 29.82 & 20.29 & 15.90 \\
\hline
Ours (500-500) & \textbf{55.69} & \textbf{57.55} & \textbf{42.22} & \textbf{53.29} & \textbf{45.55} & \textbf{39.65} & \textbf{34.05} & \textbf{30.52} & \textbf{20.59} & \textbf{16.46}  \\
\hline
\end{tabular}
\caption{ Evaluations of ChatGLM with different sizes of candidates and training samples on CANARD dataset. ``500-100'' denotes experiments with 500 candidates and 100 training samples, and so on.}
\label{size}
\end{table*}

\begin{table*}
\centering
\begin{tabular}{l|c|c|c|c|c|c|c|c|c|c}
\hline
 \multirow{2}{*}{Model}  & \multicolumn{3}{c|}{ROUGE} & \multicolumn{4}{c|}{BLEU} & \multicolumn{3}{c}{F-score} \\
 \cline{2-11}
 &  \textbf{RL} &  \textbf{R1} & \textbf{R2} &  \textbf{B1} & \textbf{B2}  & \textbf{B3} & \textbf{B4} & \textbf{F1} & \textbf{F2}  & \textbf{F3}  \\
\hline
EPR-3 &  53.91 &  55.92 &  40.14 & 51.44 & 43.47 & 37.45 & 31.86 & 29.15 & 20.05 & 15.83   \\
Ours-3 & 54.90 & 56.01 & 41.22 & 49.45 & 42.80 & 37.57 & 32.53 & 29.60 & 20.15 & 16.03  \\
\hline
EPR-4  &  54.12 &  56.27 &  40.45 & 51.36 & 43.40 & 37.37 & 31.76 & 29.64 & 20.01 & 15.59   \\
Ours-4 & 55.11 & 57.29 & 41.57 & 52.71 & 44.81 & 38.84 & 33.37 & \textbf{30.92} & \textbf{21.12} & \textbf{16.75}  \\
\hline
EPR-5 &  54.08 & 56.26 & 40.23 & 51.59 & 43.54 & 37.51 & 31.91 & 29.59 & 19.96 & 15.62    \\
Ours-5 & \textbf{55.69} & \textbf{57.55} & \textbf{42.22} & \textbf{53.29} & \textbf{45.55} & \textbf{39.65} & \textbf{34.05} & 29.52 & 20.59 & 16.46  \\
\hline
\end{tabular}
\caption{ Evaluations of ChatGLM with different numbers of examples in the demonstration on CANARD dataset. ``EPR-3'' denotes the baseline EPR with 3 examples in the demonstration.}
\label{shot}
\end{table*}

\begin{table}
\centering
\begin{tabular}{l|c|c|c}
\hline
 Model &  F1 & F2 & F3 \\
\hline
QUEEN  & 20.33 & 13.25 & 11.59  \\
Ours  & 29.52 & 20.59 & 16.46  \\
\hline
\end{tabular}
\caption{ Comparing with SFT model in few-shot setting.}
\label{sft}
\end{table}

\begin{table*}
\centering
\begin{tabular}{l|c|c|c|c|c|c}
\hline
 \multirow{2}{*}{Model}  & \multicolumn{3}{c|}{Incomplete} & \multicolumn{3}{c}{Rewritten} \\
 \cline{2-7}
 &  \textbf{Length} &  \textbf{POS} & \textbf{Chunk}  &  \textbf{Length} &  \textbf{POS} & \textbf{Chunk}  \\
\hline
BSR &  8.55 &  6.29 &  2.33 & 11.91 & 7.78 & 3.45  \\
\hline
Ours & \textbf{10.81} & \textbf{6.99} & \textbf{3.23} & \textbf{13.36} & \textbf{7.95} & \textbf{4.03} \\
\hline
\end{tabular}
\caption{ Complexity and abundance of example selected. ``Incomplete'' denotes the metrics of incomplete utterance.``Rewritten'' denotes the metrics of rewritten utterance, 
``Length'', ``POS'', and ``Chunk'' denote utterance length, number of POS types and number of text chunks respectively. 
}
\label{complexity}
\end{table*}

\begin{table*}
\centering
\begin{tabular}{l|c|c|c|c|c|c|c|c|c|c}
\hline
 \multirow{3}{*}{Model}  & \multicolumn{3}{c|}{ROUGE} & \multicolumn{4}{c|}{BLEU} & \multicolumn{3}{c}{F-score} \\
 \cline{2-11}
 &  \textbf{RL} &  \textbf{R1} & \textbf{R2} &  \textbf{B1} & \textbf{B2}  & \textbf{B3} & \textbf{B4} & \textbf{F1} & \textbf{F2}  & \textbf{F3}  \\
\hline
Length & 50.84 &  52.13 &  37.17 & 46.76  & 39.41 & 33.89 & 28.67 & 19.95 & 14.46 & 11.99   \\
POS & 52.17 &  53.77 &  38.69 & 47.92  & 40.44 & 34.80 & 29.46 & 23.89 & 17.09 & 13.92   \\
Chunk & 53.68 &  55.20 &  40.34 & 52.09  & 44.68 & 38.86 & 33.29 & 24.70 & 18.15 & 14.90   \\
\hline
Ours & \textbf{55.69} & \textbf{57.55} & \textbf{42.22} & \textbf{53.29} & \textbf{45.55} & \textbf{39.65} & \textbf{34.05} & \textbf{29.52} & \textbf{20.59} & \textbf{16.46}  \\
\hline
\end{tabular}
\caption{ ICL Evaluations of ChatGLM with 5-shot demonstrations on CANARD dataset.}
\label{chunk}
\end{table*}

\begin{table*}
\centering
\begin{tabular}{l|c|c|c|c|c|c|c|c|c|c}
\hline
 \multirow{3}{*}{Model}  & \multicolumn{3}{c|}{ROUGE} & \multicolumn{4}{c|}{BLEU} & \multicolumn{3}{c}{F-score} \\
 \cline{2-11}
 &  \textbf{RL} &  \textbf{R1} & \textbf{R2} &  \textbf{B1} & \textbf{B2}  & \textbf{B3} & \textbf{B4} & \textbf{F1} & \textbf{F2}  & \textbf{F3}  \\
\hline
Random &  78.07 &  79.34 &  66.50 & 69.05  & 64.15 & 60.26 & 56.35 & 52.86 & 43.49 & 38.14   \\
\hline
BSR & 80.13 & 81.24 & 68.78 & 73.54 & 67.37 & 65.42 & 59.78 & 56.16  & 48.36 & 44.89 \\
\hline
Ours & \textbf{82.77} & \textbf{84.04} & \textbf{70.37} & \textbf{79.07} & \textbf{73.81} & \textbf{69.56} &  \textbf{65.47} & \textbf{62.64} & \textbf{52.61} & \textbf{47.30}  \\
\hline
\end{tabular}
\caption{ ICL Evaluations of gpt3.5 with 5-shot demonstrations on TASK dataset.}
\label{gpt3.5}
\end{table*}

\begin{table}
\centering
\begin{tabular}{l|c|c|c|c|c}
\hline
 \multirow{3}{*}{Model}  & \multicolumn{2}{c|}{ROUGE} & \multicolumn{3}{c}{BLEU}  \\
 \cline{2-6}
 &  \textbf{R1} & \textbf{R2} &  \textbf{B1} & \textbf{B2}  & \textbf{B3}  \\
\hline
BSR &  51.6 & 37.8 & 35.2 & 29.6 & 26.4   \\
Ours & 57.9 & 45.1 & 43.0 & 38.3 & 35.2  \\
Ours(r)  &  57.6 &  44.5 & 43.3  & 38.6 & 34.9   \\
\hline
\end{tabular}
\caption{ ICL Evaluations of example orders with ChatGLM on TASK dataset.}
\label{order}
\end{table}

\subsection{Experimental Details}
Following \citet{liu2022few} and \citet{rubin2021learning} we utilize SentenceBERT \cite{reimers2019sentence} as LM selector for English datasets and bert-base-Chinese for Chinese datasets. The sizes of candidate example set $C$ and training set $T$ are 500. The learning rate is set to be 1e-5. The task instruction in section \ref{format} is designed as ``Rewrite an incomplete utterance into an utterance which is semantically equivalent but self-contained to be understood without context. The sentence structure and expression should be consistent.'' for English dataset and its translated version for Chinese dataset.

\subsection{Results}
In Table \ref{canard}, our approach outperforms all the baselines by about 1.2-1.7 ROUGE score, 1.3-2.5 BLEU score, 0.4 F2 score and 0.6 F3 score in CANARD dataset. It demonstrates the efficiency of directly utilizing feedback by LLM and RL policy gradient to train the LM selector. With the examples selected by our method, the performances of LLM are significantly improved. Our model not only improves the overall quality of utterance rewritten, but also captures the important words from the context.

Compared with Random selecting examples, both BM25 and KATE baselines derive better performance. It shows the textual similarity captured by sparse retrieval like BM25 and the semantic similarity captured by KATE can help select better examples to prompt the LLM for IUR task.
EPR and BSR show an overall better performance compared with the sole sparse or dense retrieval methods. It demonstrates the combination of sparse or dense retrieval can capture both the textual and semantic similarity between candidate examples and the test case. The examples selected are better prompts to LLM for IUR task.
Compared with EPR, BSR behaves a better performance across different metrics. BM25 behaves as the effective supervision for the LM retriever and releasing the constraint can improve the example selection furthermore.

In TASK dataset, our approach outperforms all the baselines by about 
6.3-7.5 ROUGE score and 7.6-8.2 BLEU score. Compared with CANARD dataset, TASK contains more complex and diverse dialogue topics. It demonstrates the efficiency of utilizing direct LLM feedback to train the LM retriever and select examples to prompt the LLM for IUR task.
In REWRTTE dataset, our approach outperforms all the baselines by about 0.6 ROUGE and BLEU score, which shows our stable efficiency with different languages.

\section{Analysis}
In this part, first we explore our performance with different sizes of candidate set and training set. Then we probe the effect of different numbers of examples in demonstrations. Furthermore, we compare our approach with the SFT method in few shot setting. Finally, we explore the reason why examples selected by our approach can improve the analogy ability of LLM.

\subsection{Different Sizes of Candidates and Training Samples}
In Table \ref{canard}, the number of candidates $\#C$ and training samples $\#T$ are 500. In this part, we do further experiments with $\mathbf{\#C}=100, \mathbf{\#T}=100$; $\mathbf{\#C}=100, \mathbf{\#T}=500$; $\mathbf{\#C}=500, \mathbf{\#T}=100$ and freeze other hyperparameters respectively. In Table \ref{size}, generally, with more candidates and training samples, our approach will select better examples for the demonstration. With 100 candidates and 100 training samples, our approach beats random selection by about 2.0 ROUGE score, 2.5 BLEU score and 2.0 F-score. It is also comparable to the competitive baseline BSR with 500 candidates and training samples. It shows the efficiency to utilize direct LLM feedback to train the LM retriever. 
Compared with setting 100-500, our approach outperforms by about 0.7 ROUGE score, 0.8 BLEU score and 0.9 F-score. It demonstrates our ability to select better examples to improve the analogy ability of LLM With more candidates.
Compared with setting 500-100, our approach outperforms by about 1.4 ROUGE score, 1.9 BLEU score and 0.2 F-score. It shows more training samples improve the selection of our approach from the candidate set for IUR task.

\subsection{Different Number of Examples in Demonstration}
In Table \ref{canard}, we conduct the experiments with 5-shot demonstrations. In this part, we do further experiments with 3-shot and 4-shot demonstrations. In Table \ref{shot}, generally with more examples in the demonstration, our approach improves the ICL performance of LLM for IUR task. Especially, with more demonstration examples, our approach derives more improvement compared with the competitive baseline EPR. It demonstrates the efficiency of our RLS by directly utilizing LLM feedback to train the LM retriever and improve the ICL performance of LLM.

\subsection{Comparing with SFT Model}
In this part, we compare our approach with the existing state-of-the-art QUEEN \cite{liu2020incomplete} in IUR field. QUEEN tackles IUR task by finetuning PTM \cite{devlin2018bert} and constructing the word edit matrix. Different from QUEEN , our approach utilizes LM as a proxy to select appropriate examples and parameters of the answer generator ChatGLM are fixed. To keep a fair comparison, we assign the candidate set and training set in our approach as the training set of QUEEN. 
In Table \ref{sft}, our approach QUEEN by 9.2 F1 score, 7.3 F2 score and 4.9 F3 score. F-score concentrate on the words from the context, which are argued to be harder to copy \cite{pan2019improving}. It shows our efficiency to capture important words from the context to rewrite the incomplete utterance. Considering our RLS approach does not depend on the LLM server (ChatGLM in our experiments), it is promising for our approach to derive better results for IUR task with stronger LLM.

\subsection{What Examples are Appropriate for IUR}
In this part, we explore the reason why examples selected by our approach serve as better demonstrations for the LLM to solve IUR task.

We assume that other than the textual and semantic similarity of examples with the current test case, the complexity and abundance of examples in the demonstrations also matter for IUR task. 
We evaluate the complexity and abundance of examples by three metrics of incomplete utterance and rewritten utterance: 1. the length of utterance; 2. the number of Part Of Speech (POS) tagging types \cite{kumawat2015pos}. 3. the number of text chunks \cite{ramshaw1999text}. We assume the examples are more complex and abundant with longer utterances, more POS tag types and more text chunks. In practice, we adopt SpaCy to do POS tagging and text chunking.
In Table \ref{complexity}, the examples selected by our policy-based RL framework have longer utterances, more POS types and more text chunks though without explicit assignment, We argue that the complexity and abundance of examples in the demonstration are important to improve the analogy ability of LLM.

What if we select the examples only by metrics of complexity and abundance?
To address this issue, we select examples from the candidate set to construct the demonstration by metrics of the length of utterance, the number of POS types, the number of text chunks respectively and conduct the experiments on CANARD dataset.
In Table \ref{chunk}, if selecting the examples only by the metric of number of text chunks, the ICL results will drop by about 2.1 ROUGE score, 0.9 BLEU score, 2.9 F-score on CANARD dataset. It will drop more if selecting the examples only by the metric of utterance length or POS types. It shows only selecting examples by complexity and abundance, the LLM performance will not necessarily improve. Balancing the abundance and the similarity with the test case of examples is important to improve the ICL ability of LLM. Our approach attains the balance of complexity and abundance, as well as semantic similarity with the test case without explicit assignment.

\subsection{Efficiency of Larger LLM}
In this part, we conduct the experiments by applying our approach with gpt3.5 in TASK dataset. In Table \ref{gpt3.5}, our RLS outperforms random selecting examples by about 4.4 ROUGE score, 9.5 BLEU score and 9.4 F-score in TASK dataset. It demonstrates the efficiency of directly utilizing LLM feedback to train the LM retriever with policy-based RL and improve the analogy ability of larger LLM. With the emergence of larger LLM, it is promising for our approach to select appropriate examples to improve the ICL performance of LLM.

\subsection{Order of Examples}
In this part, we probe the effect of the order of examples in the demonstration. We arrange examples in the demonstration by the order of sampling with Eq.\ref{prob} or the reverse order and conduct the experiments on TASK dataset. Table \ref{order} shows if arranging the examples by the reverse order of sampling, our approach shows comparable performance . It demonstrates the stable efficiency of our directly selecting the examples that can improve the analogy ability of LLM by policy gradient.

\section{Conclusion}
Existing example selection methods fail to directly utilize the LLM feedback to choose appropriate examples in the demonstration. We propose a novel and effective example selection framework by directly adopting the LLM feedback to train the LM selector with policy-based RL. Our approach significantly improves the ICL performance of LLM for IUR tasks. 

\section*{Limitations}
we propose our novel and effective example selection framework by directly adopting the LLM feedback to train the LM selector with policy-based RL and demonstrate our efficiency on three benchmark datasets in this field. Though only utilizing the LLM ChatGLM-7B and gpt3.5, we will conduct experiments with diverse kinds of LLMs in future research.

\bibliography{anthology,custom}

\begin{thebibliography}{33}
\expandafter\ifx\csname natexlab\endcsname\relax\def\natexlab#1{#1}\fi

\bibitem[{Brown et~al.(2020)Brown, Mann, Ryder, Subbiah, Kaplan, Dhariwal, Neelakantan, Shyam, Sastry, Askell et~al.}]{brown2020language}
Tom Brown, Benjamin Mann, Nick Ryder, Melanie Subbiah, Jared~D Kaplan, Prafulla Dhariwal, Arvind Neelakantan, Pranav Shyam, Girish Sastry, Amanda Askell, et~al. 2020.
\newblock Language models are few-shot learners.
\newblock \emph{Advances in neural information processing systems}, 33:1877--1901.

\bibitem[{Choi et~al.(2018)Choi, He, Iyyer, Yatskar, Yih, Choi, Liang, and Zettlemoyer}]{choi2018quac}
Eunsol Choi, He~He, Mohit Iyyer, Mark Yatskar, Wen-tau Yih, Yejin Choi, Percy Liang, and Luke Zettlemoyer. 2018.
\newblock Quac: Question answering in context.
\newblock \emph{arXiv preprint arXiv:1808.07036}.

\bibitem[{Chowdhery et~al.(2022)Chowdhery, Narang, Devlin, Bosma, Mishra, Roberts, Barham, Chung, Sutton, Gehrmann et~al.}]{chowdhery2022palm}
Aakanksha Chowdhery, Sharan Narang, Jacob Devlin, Maarten Bosma, Gaurav Mishra, Adam Roberts, Paul Barham, Hyung~Won Chung, Charles Sutton, Sebastian Gehrmann, et~al. 2022.
\newblock Palm: Scaling language modeling with pathways.
\newblock \emph{arXiv preprint arXiv:2204.02311}.

\bibitem[{Devlin et~al.(2018)Devlin, Chang, Lee, and Toutanova}]{devlin2018bert}
Jacob Devlin, Ming-Wei Chang, Kenton Lee, and Kristina Toutanova. 2018.
\newblock Bert: Pre-training of deep bidirectional transformers for language understanding.
\newblock \emph{arXiv preprint arXiv:1810.04805}.

\bibitem[{Dong et~al.(2022)Dong, Li, Dai, Zheng, Wu, Chang, Sun, Xu, and Sui}]{dong2022survey}
Qingxiu Dong, Lei Li, Damai Dai, Ce~Zheng, Zhiyong Wu, Baobao Chang, Xu~Sun, Jingjing Xu, and Zhifang Sui. 2022.
\newblock A survey for in-context learning.
\newblock \emph{arXiv preprint arXiv:2301.00234}.

\bibitem[{Elgohary et~al.(2019)Elgohary, Peskov, and Boyd-Graber}]{elgohary2019can}
Ahmed Elgohary, Denis Peskov, and Jordan Boyd-Graber. 2019.
\newblock Can you unpack that? learning to rewrite questions-in-context.
\newblock \emph{Can You Unpack That? Learning to Rewrite Questions-in-Context}.

\bibitem[{Gonen et~al.(2022)Gonen, Iyer, Blevins, Smith, and Zettlemoyer}]{gonen2022demystifying}
Hila Gonen, Srini Iyer, Terra Blevins, Noah~A Smith, and Luke Zettlemoyer. 2022.
\newblock Demystifying prompts in language models via perplexity estimation.
\newblock \emph{arXiv preprint arXiv:2212.04037}.

\bibitem[{Gupta et~al.(2023)Gupta, Singh, and Gardner}]{gupta2023coverage}
Shivanshu Gupta, Sameer Singh, and Matt Gardner. 2023.
\newblock Coverage-based example selection for in-context learning.
\newblock \emph{arXiv preprint arXiv:2305.14907}.

\bibitem[{Hao et~al.(2021)Hao, Song, Wang, Xu, Tu, and Yu}]{hao2021rast}
Jie Hao, Linfeng Song, Liwei Wang, Kun Xu, Zhaopeng Tu, and Dong Yu. 2021.
\newblock Rast: Domain-robust dialogue rewriting as sequence tagging.
\newblock In \emph{Proceedings of the 2021 Conference on Empirical Methods in Natural Language Processing}, pages 4913--4924.

\bibitem[{Huang et~al.(2021)Huang, Li, Zou, and Zhang}]{huang2021sarg}
Mengzuo Huang, Feng Li, Wuhe Zou, and Weidong Zhang. 2021.
\newblock Sarg: A novel semi autoregressive generator for multi-turn incomplete utterance restoration.
\newblock In \emph{Proceedings of the AAAI Conference on Artificial Intelligence}, volume~35, pages 13055--13063.

\bibitem[{Inoue et~al.(2022)Inoue, Liu, Son, and Nguyen}]{inoue2022enhance}
Shumpei Inoue, Tsungwei Liu, Nguyen~Hong Son, and Minh-Tien Nguyen. 2022.
\newblock Enhance incomplete utterance restoration by joint learning token extraction and text generation.
\newblock \emph{arXiv preprint arXiv:2204.03958}.

\bibitem[{Kim et~al.(2022)Kim, Cho, Kim, Kim, Yoo, and Lee}]{kim2022self}
Hyuhng~Joon Kim, Hyunsoo Cho, Junyeob Kim, Taeuk Kim, Kang~Min Yoo, and Sang-goo Lee. 2022.
\newblock Self-generated in-context learning: Leveraging auto-regressive language models as a demonstration generator.
\newblock \emph{arXiv preprint arXiv:2206.08082}.

\bibitem[{Kumawat and Jain(2015)}]{kumawat2015pos}
Deepika Kumawat and Vinesh Jain. 2015.
\newblock Pos tagging approaches: A comparison.
\newblock \emph{International Journal of Computer Applications}, 118(6).

\bibitem[{Levy et~al.(2022)Levy, Bogin, and Berant}]{levy2022diverse}
Itay Levy, Ben Bogin, and Jonathan Berant. 2022.
\newblock Diverse demonstrations improve in-context compositional generalization.
\newblock \emph{arXiv preprint arXiv:2212.06800}.

\bibitem[{Lin(2004)}]{lin2004rouge}
Chin-Yew Lin. 2004.
\newblock Rouge: A package for automatic evaluation of summaries.
\newblock In \emph{Text summarization branches out}, pages 74--81.

\bibitem[{Liu et~al.(2022)Liu, Tam, Muqeeth, Mohta, Huang, Bansal, and Raffel}]{liu2022few}
Haokun Liu, Derek Tam, Mohammed Muqeeth, Jay Mohta, Tenghao Huang, Mohit Bansal, and Colin~A Raffel. 2022.
\newblock Few-shot parameter-efficient fine-tuning is better and cheaper than in-context learning.
\newblock \emph{Advances in Neural Information Processing Systems}, 35:1950--1965.

\bibitem[{Liu et~al.(2021)Liu, Shen, Zhang, Dolan, Carin, and Chen}]{liu2021makes}
Jiachang Liu, Dinghan Shen, Yizhe Zhang, Bill Dolan, Lawrence Carin, and Weizhu Chen. 2021.
\newblock What makes good in-context examples for gpt-$3 $?
\newblock \emph{arXiv preprint arXiv:2101.06804}.

\bibitem[{Liu et~al.(2020)Liu, Chen, Lou, Zhou, and Zhang}]{liu2020incomplete}
Qian Liu, Bei Chen, Jian-Guang Lou, Bin Zhou, and Dongmei Zhang. 2020.
\newblock Incomplete utterance rewriting as semantic segmentation.
\newblock \emph{arXiv preprint arXiv:2009.13166}.

\bibitem[{Pan et~al.(2019)Pan, Bai, Wang, Zhou, and Liu}]{pan2019improving}
Zhufeng Pan, Kun Bai, Yan Wang, Lianqiang Zhou, and Xiaojiang Liu. 2019.
\newblock Improving open-domain dialogue systems via multi-turn incomplete utterance restoration.
\newblock In \emph{Proceedings of the 2019 Conference on Empirical Methods in Natural Language Processing and the 9th International Joint Conference on Natural Language Processing (EMNLP-IJCNLP)}, pages 1824--1833.

\bibitem[{Papineni et~al.(2002)Papineni, Roukos, Ward, and Zhu}]{papineni2002bleu}
Kishore Papineni, Salim Roukos, Todd Ward, and Wei-Jing Zhu. 2002.
\newblock Bleu: a method for automatic evaluation of machine translation.
\newblock In \emph{Proceedings of the 40th annual meeting of the Association for Computational Linguistics}, pages 311--318.

\bibitem[{Quan et~al.(2019)Quan, Xiong, Webber, and Hu}]{quan2019gecor}
Jun Quan, Deyi Xiong, Bonnie Webber, and Changjian Hu. 2019.
\newblock Gecor: An end-to-end generative ellipsis and co-reference resolution model for task-oriented dialogue.
\newblock \emph{arXiv preprint arXiv:1909.12086}.

\bibitem[{Ramshaw and Marcus(1999)}]{ramshaw1999text}
Lance~A Ramshaw and Mitchell~P Marcus. 1999.
\newblock Text chunking using transformation-based learning.
\newblock \emph{Natural language processing using very large corpora}, pages 157--176.

\bibitem[{Reddy et~al.(2019)Reddy, Chen, and Manning}]{reddy2019coqa}
Siva Reddy, Danqi Chen, and Christopher~D Manning. 2019.
\newblock Coqa: A conversational question answering challenge.
\newblock \emph{Transactions of the Association for Computational Linguistics}, 7:249--266.

\bibitem[{Reimers and Gurevych(2019)}]{reimers2019sentence}
Nils Reimers and Iryna Gurevych. 2019.
\newblock Sentence-bert: Sentence embeddings using siamese bert-networks.
\newblock \emph{arXiv preprint arXiv:1908.10084}.

\bibitem[{Robertson et~al.(2009)Robertson, Zaragoza et~al.}]{robertson2009probabilistic}
Stephen Robertson, Hugo Zaragoza, et~al. 2009.
\newblock The probabilistic relevance framework: Bm25 and beyond.
\newblock \emph{Foundations and Trends{\textregistered} in Information Retrieval}, 3(4):333--389.

\bibitem[{Rubin et~al.(2021)Rubin, Herzig, and Berant}]{rubin2021learning}
Ohad Rubin, Jonathan Herzig, and Jonathan Berant. 2021.
\newblock Learning to retrieve prompts for in-context learning.
\newblock \emph{arXiv preprint arXiv:2112.08633}.

\bibitem[{Sorensen et~al.(2022)Sorensen, Robinson, Rytting, Shaw, Rogers, Delorey, Khalil, Fulda, and Wingate}]{sorensen2022information}
Taylor Sorensen, Joshua Robinson, Christopher~Michael Rytting, Alexander~Glenn Shaw, Kyle~Jeffrey Rogers, Alexia~Pauline Delorey, Mahmoud Khalil, Nancy Fulda, and David Wingate. 2022.
\newblock An information-theoretic approach to prompt engineering without ground truth labels.
\newblock \emph{arXiv preprint arXiv:2203.11364}.

\bibitem[{Su et~al.(2019)Su, Shen, Zhang, Sun, Hu, Niu, and Zhou}]{su2019improving}
Hui Su, Xiaoyu Shen, Rongzhi Zhang, Fei Sun, Pengwei Hu, Cheng Niu, and Jie Zhou. 2019.
\newblock Improving multi-turn dialogue modelling with utterance rewriter.
\newblock \emph{arXiv preprint arXiv:1906.07004}.

\bibitem[{Sun et~al.(2019)Sun, Yu, Chen, Yu, Choi, and Cardie}]{sun2019dream}
Kai Sun, Dian Yu, Jianshu Chen, Dong Yu, Yejin Choi, and Claire Cardie. 2019.
\newblock Dream: A challenge data set and models for dialogue-based reading comprehension.
\newblock \emph{Transactions of the Association for Computational Linguistics}, 7:217--231.

\bibitem[{Touvron et~al.(2023)Touvron, Lavril, Izacard, Martinet, Lachaux, Lacroix, Rozi{\`e}re, Goyal, Hambro, Azhar et~al.}]{touvron2023llama}
Hugo Touvron, Thibaut Lavril, Gautier Izacard, Xavier Martinet, Marie-Anne Lachaux, Timoth{\'e}e Lacroix, Baptiste Rozi{\`e}re, Naman Goyal, Eric Hambro, Faisal Azhar, et~al. 2023.
\newblock Llama: Open and efficient foundation language models.
\newblock \emph{arXiv preprint arXiv:2302.13971}.

\bibitem[{Vasiliev(2020)}]{vasiliev2020natural}
Yuli Vasiliev. 2020.
\newblock \emph{Natural language processing with Python and spaCy: A practical introduction}.
\newblock No Starch Press.

\bibitem[{Zhang et~al.(2022)Zhang, Li, Wang, Cheng, and Xiao}]{zhang2022self}
Yong Zhang, Zhitao Li, Jianzong Wang, Ning Cheng, and Jing Xiao. 2022.
\newblock Self-attention for incomplete utterance rewriting.
\newblock In \emph{ICASSP 2022-2022 IEEE International Conference on Acoustics, Speech and Signal Processing (ICASSP)}, pages 8047--8051. IEEE.

\bibitem[{Zhao et~al.(2021)Zhao, Wallace, Feng, Klein, and Singh}]{zhao2021calibrate}
Zihao Zhao, Eric Wallace, Shi Feng, Dan Klein, and Sameer Singh. 2021.
\newblock Calibrate before use: Improving few-shot performance of language models.
\newblock In \emph{International Conference on Machine Learning}, pages 12697--12706. PMLR.

\end{thebibliography}
\bibliographystyle{acl_natbib}

\appendix

\section{Example Appendix}
\label{sec:appendix}

This is an appendix.

\end{document}